\documentclass[]{article}
\usepackage{amsmath, amssymb}
\usepackage{graphicx, subcaption}
\usepackage{float}
\usepackage{hyperref}
\hypersetup{
	colorlinks=true,
	linkcolor=blue,
	filecolor=magenta,      
	urlcolor=cyan,
}
\usepackage{tikz}

\usepackage{booktabs}
\usepackage{array}
\usepackage{authblk}

\title{Semisupervised Adversarial Neural Networks for Cyber Security Transfer Learning}
\author{Casey Kneale\footnote{Corresponding author: Casey Kneale, ckneale@ccri.com}, Kolia Sadeghi}
\affil{Commonwealth Computer Research, Inc. \\ Sachem Place, Charlottesville, VA}

\begin{document}


\maketitle

\begin{abstract}
	On the path to establishing a global cybersecurity framework where each enterprise shares information about malicious behavior, an important question arises. How can a machine learning representation characterizing a cyber attack on one network be used to detect similar attacks on other enterprise networks if each networks has wildly different distributions of benign and malicious traffic? We address this issue by comparing the results of naively transferring a model across network domains and using CORrelation ALignment, to our novel adversarial Siamese neural network. Our proposed model learns attack representations that are more invariant to each network's particularities via an adversarial approach. It uses a simple ranking loss that prioritizes the labeling of the most egregious malicious events correctly over average accuracy. This is appropriate for driving an alert triage workflow wherein an analyst only has time to inspect the top few events ranked highest by the model. In terms of accuracy, the other approaches fail completely to detect any malicious events when models were trained on one dataset are evaluated on another for the first 100 events. While, the method presented here retrieves sizable proportions of malicious events, at the expense of some training instabilities due in adversarial modeling. We evaluate these approaches using 2 publicly available networking datasets, and suggest areas for future research.
\end{abstract}

\section{Introduction}
	The large differences in distributions of data from networks operating under different conditions introduce challenges for frameworks aiming to learn attack signatures globally across all networks. Global analysis involves the sharing of models or information across one or more enterprises. If two or more enterprises have data that are statistically similar, then the application of a single model from either domain can be applied to the other typically with minor or no loss in performance. When the domains are different, the context of a model then depends upon the domain it was derived from. Ideally, the characterization of cybersecurity threats with invariant representations would allow for universal detection across enterprise networks. In this work we specifically address the problem of context dependence for purposes of shared threat identification across two exemplary domains. 

	We first show that network data collected from separate networks, enterprises, or at different times can be statistically different. These differences are likely the result of different enterprises using different sets of networked services, technological advances over time (Moore's Law) \cite{Mooreslaw}, new attack vectors, and/or different trends in end-user behavior. For example, one could anticipate that network data that has been collected from a corporate network during work hours would feature comparatively less streaming/social media, but potentially more SMTP traffic than network data obtained from a university on a weekend. These real world phenomena make it difficult to utilize academic datasets, or data from other enterprises in global models.
	
	What does an analyst want to experience when they bring a new system online? Many working analysts make use of ticketing systems, where end users, or other analysts, report problems in the form of a \textit{ticket}. The recorded tickers are later read and addressed via human intervention. An effective ticketing system places tasks/threats that require priority to the top of a list so they are addressed in a timely manner\cite{Tickets}. This type of security workflow is often referred to as triage \cite{Triage}. Triage reduces the time spent examining normal/mislabeled network traffic, and large volumes of low priority risks, when another incident with greater importance is reported. 
	
	In practice it is expensive for each new enterprise in a global framework to create their own labeled datasets from scratch. This is because classification and triage models can require hundreds, if not thousands, of malicious examples. An approach to solving this problem found in the field of machine learning is to make models context independent in the presence of interdomain variance, so they can be transferred across domains\cite{DANN,CORAL}. The goal of this study is to demonstrate a proof of concept for how preexisting information from a labeled network, can immediately be utilized for triage on a network with different data distributions for which no events have been labeled. 
	
	For proof of concept we show that the transfer of a triage model which targets Denial of Service (DoS) attacks between either the UNSW NB-15 or the CICIDS2017 datasets can be performed. DoS attacks were selected as a proof of concept but also due to necessity; there is a concerning lack of publicly available cyber security data which has similar class labels. Despite that it may seem trivial to transfer a DoS detector because DoS has historically had a similar attack vector across time \cite{DosReview} we show that minimal effort approaches fail to produce useful transfers with internetwork variances present in the aforementioned data. 
	
	We demonstrate the efficacy of two approaches from the field of machine learning for transductively transferring\cite{TransferLearnReview} a triage model from one academic dataset to another. The transfer methods we describe are performed in either an unsupervised or semi-supervised manner, meaning that the class labels of the domain receiving label information from the other is not accounted for in the modeling. We also present a novel approach to assessing triage models called Rolling TopN accuracy, and some useful metrics for training the models employed. Although the effort presented here is preliminary, and thereby limited in scope, we show that for triage purposes, adversarial neural networks offer a promising route to challenges inherent with domain transfers in context dependent settings. 

\subsection{Related Work}
Robust Representation for Domain Adaptation in Network Security \cite{RobustRep}
Machine Learning in Cyber-Security Problems, Challenges and DataSets \cite{CyberDomainAdapt}
Transfer learning for detecting unknown network attacks \cite{CHeTL}

\section{Data}
All data was preprocessed in the following way: quantiles for each variable were clamped to be $\geq$ 0.001 and $\leq$ 0.999 before being range/min-max scaled. This preprocessing was used to trim some of the long tails of the features employed and place the data into a reasonable range for weight optimization via back propagation. It is important to state that this preprocessing was performed independently across domains, and only used information available from the respective training sets when applied to hold out sets. Where supervised learning was employed, only the first 80,000 samples of the dataset were considered to be a training set, and the remaining samples to be the test set. This was done to ensure a realistic scenario where a new network is being brought online and temporal order is maintained.

\subsection{UNSW NB-15}
The UNSW NB-15 dataset was collected in 2015 by a semisynthetic network \cite{UNSW}. The network was composed of artificial traffic generated by the IXIA perfect storm tool and traffic generated by several human users across a fire-wall. It features 9 attack categories, and is available in several formats (PCAP, BRO, CSV). The CSV datasets were used for convenience. The class balance in the for DoS to normal traffic is approximately 1:19 and 1:13 for the train and test sets respectively. Please note, the other 8 attack classes in this data were pooled into a 'normal' traffic label to mimic the conditions of having mostly unlabeled network data.

\subsection{CICIDS2017}
The CICIDS2017 dataset was collected 2 years after the UNSW NB-15 data. The collection took place for a week. On each day attacks such as: Brute Force FTP, Brute Force SSH, DoS, Heartbleed, Web Attack, Infiltration, Botnet and DDoS were conducted. For this study we used the DoS attacks from the ``Wednesday" and Distributed DoS (DDoS) attacks in the ``Friday" subsets. A detailed list of the DoS and DDoS attacks present on these subsets is provided in table. The prevalence of DoS network activity relative to normal activity on the Wednesday subset was 1:1.76 and the Friday subset had slightly more than 1:1. For more information about the network topology and experimental design, please see the original reference \cite{CICIDS}.

\begin{table}[H]
	\centering
	\begin{tabular}{ c | c }
		\toprule
		Wednesday & Friday \\
		\midrule
		DoS slowloris, DoS Slowhttptest, & DDoS LOIT, Botnet ARES \\ 
		DoS Hulk, DoS GoldenEye  & - \\
	\end{tabular}
	\caption{Different DoS and DDoS attacks conducted on Wednesday and Friday of the CICIDS2017 dataset. }
\end{table}

\subsection{ Feature Engineering }
In their native forms only a few features between both datasets are shared (IE: UNSW NB-15's "dur" and CICIDS2017's "Flow\_Duration"). The following 7 column labels available form the UNSW NB-15 CSV dataset: dur, spkts, dpkts, load, rate, sinpkt, and dinpkt, were calculated/obtained from available features in the CICIDS2017 dataset. Aside from dimensional analysis no further engineering was performed.

\section{Methods}

\subsection{ CORAL }
CORrelation ALignment, or CORAL, is a simple domain adaptation method often used in the field of transfer learning. The objective function used in CORAL minimizes the Frobenius norm of the differences between two domains' ($A$ \& $B$) covariance matrices($C_A$ \& $C_B$) by solving for some transfer matrix $T$ such that $||T^TC_AT - C_B||^F$ is minimal. As suggested in the original publication the transfer matrix can be obtained via whitening the source covariance matrix and recoloring with the target covariance in 4 lines of linear algebra supported code. This method is attractive because it is independent of sample correspondences across domains because it seeks to map the second order statistics of the two domains\cite{CORAL}. The alignment of similar samples does not need to be performed.

\subsection{ Adversarial Neural Networks }
Adversarial neural networks (AN) are composed of at least one primary network trained to perform the task at hand (Ie: classification / ranking) and at least one adversarial network which challenges the primary network. The sub-network or external adversarial network is typically trained on an auxiliary but related task to generalize the results which would otherwise be obtained from being trained solely on the primary task. This type of training can be seen as a zero sum game where one network is poised to minimize it's modeling loss ($L_P$) but the other is poised to maximize the error in the primary task ($L_A$). Loss functions of this variety have the following general form:
\begin{equation} \label{eq:1}
	Loss = max( min( L_P ) - L_A )
\end{equation}

Many AN's, especially generative adversarial networks, are composed of two models which do not share weights\cite{GANS}. In our hands we have found those models difficult to train. Instead, a trending approach commonly used in transfer learning tasks is to share weights with-in the model and build separate outputs or \textit{heads} for the primary and adversarial tasks \cite{DANN, USAD}. 

Our network consisted of three basic units, the first is a feed-forward neural network chain which creates an embedded representation of the input data($E$), a ListMLE \cite{ListMLE} subnetwork which ranks netflows by their label (Ie: Normal $\rightarrow$ DoS) ($L$) and an adversarial Siamese subnetwork which attempts to discriminate between CICIDS2017 and UNSW NB-15 flows($D$). This generic structure is depicted by Figure 1, and each output/head is described below.

\begin{figure}[H]
	\centering
	\includegraphics[width=0.5\textwidth]{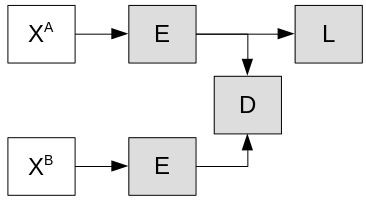}
	\caption{Adversarial Neural Network topology.}
\end{figure}

\subsection{ListMLE}
ListMLE neural networks are neural networks that do not expressly perform a classification or regression task. Instead ListMLE learns rankings of observations based on their features in a supervised manner. This goal coincides well with the idea of threat triage, in that certain malicious behaviors should take precedence over others for human diagnosis / intervention. 

In this work $N$ training observations resulted from the concatenation of DoS and normal classes (1:1). The elements of the list ranking vector($y$) were calculated for each batch using the following expression: $y_i = i/N$. This ranking implies that a value of $0.0$ corresponded to an exemplary DoS, at $0.5$ the first normal flow was observed, and at $1.0$ the final normal observation had been presented. The loss function ListMLE models employ optimizes the Plackett-Luce likelihood that the predicted ranking scores follow in the assigned order,
$$Loss_{ListMLE}(L(x),y) = - \sum_{i=1}^{N} ln \left(\frac{ exp( L( x_i ) )}{\sum_{k | y_k > y_i} { exp( L( x_k )) } } \right)$$

Here, $\{y_i\}$ are the labels that capture the desired ordering of the list. Numeric stability and performance issues can be observed when the number of items in the list are large. One way to avoid this problem is to consider only the top $n$ samples. This approach has been shown to be statistically consistent \cite{TopK}. Our model utilized only the top 150 items during training. Another advantage is that Siamese networks scale well to large numbers of classes / domains / enterprises.

\subsection{Siamese Neural Network}
Siamese neural networks \cite{OriginalSiamese} are composed of 2 copies of the same embedding network, which is trained to learn a mapping of its inputs such that the square euclidean distances ($d$) between its outputs are discriminative. In our use-case, the domains / enterprise networks that the observations came from($X_A$ \& $X_B$), are treated as labels. Pairs of observations from the same domain are assigned a ground truth label of $Y=0$, otherwise the pair's label is $Y=1$. In our work, we chose to use a contrastive loss function, with a margin hyperparameter ($m = 2.$), as is commonly done\cite{OriginalSiamese}. The margin serves to pushes the mappings of pairs of inputs from different classes to be at least $m$ distance units apart, but penalize model weights for pairs from different domains whose mappings are already more than $m$ apart from each other:
$$Loss_{contrastive} = 0.5 * ( (1 - Y) * d + Y * max(0, m - d) )$$

Siamese networks have potential advantages when compared to traditional feed forward neural network models. One advantage that can be leveraged for transfer learning, is the ability to learn new class representations with relatively few examples. This modeling regime is referred to as few shot learning, or one shot learning when only one example for a given class is available \cite{OneShot}. This is an attractive attribute for transfer learning tasks, because the labeling of samples can be comparatively expensive. Another advantage is that Siamese networks scale well to large numbers of classes / domains / enterprises.

\subsection{Adversarial Training}
The loss functions employed to train the overall adversarial network described in Equation \ref{eq:1} can be related to both the list-wise ranking($L_L$) and Siamese discriminator ($L_D$) by the following expressions:
$$L_P = L_L = Loss_{ListMLE}$$
$$L_A = L_D = Loss_{contrastive}$$
\noindent Although these two loss functions clearly describe the primary and adversarial goals of the network, they do not explicitly describe how the model was trained. The model weight updates were calculated from two separate pools of shuffled data per iteration. Although it is common for adversarial models to employ independent stochastic gradient descent/ascent updates for each subnetwork, we did not update weights until after all computations were performed. The gradients which are attributable from the $X \rightarrow E \rightarrow L$ networks were calculated by backpropagation using gradient descent as usual. 

However, the gradient updates from $X \rightarrow E \rightarrow D$ were computed via gradient ascent. The gradients from $X \rightarrow E$ had their direction inverted by multiplying by a scalar factor of $-1$. This approach is sometimes referred to as the ReverseGrad algorithm \cite{DANN}, and it is what allows the discriminator to effect the classification head's loss ($L_L$) in an adversarial way.

\subsection{Topology}
Commonly described adversarial models, such as those in the field of computer vision, are typically complicated and bolster large numbers of neurons/floating point operations. Only a simple topology was considered for the model used in our experiments despite that better network topologies likely exist,

\begin{table}[H]
	\centering
	\begin{tabular}{ c c }
		\toprule
		Subnetwork & Topology \\
		\midrule
		E & $11^\sigma$, $10^\sigma, 9^\sigma, 7^\sigma, 7^\sigma $\\ 
		L & $14^\sigma, 1^{L}$ \\
		D & $||E(X^A) - E(X^B) ||^2$ \\
	\end{tabular}
	\caption{Topology for the adversarial neural network model used in this study.}
\end{table}

\noindent The tuning of semisupervised neural networks can only be done for the supervised domain. Estimating which architecture transfers adequately across domains is limited to statistical metrics that have been described in the \textit{Training Considerations} subsection. The authors do not attempt to make claims of optimality.

\section{Results and Discussion}
\subsection{Comparison of Source Distributions}
The Kuiper test is robust nonparametric statistical test that can be used to assesses whether or not a sample and a reference distribution are equivalent \cite{KuiperStat}. To ensure that the feature space between UNSW NB-15 and CICIDS2017 were statistically different the Kuiper test was used to compare the individual feature distributions from both training subsets (80,000 examples each). The test was conducted without binning or discretizing the observations. The cumulative distribution functions were calculated by sorting the pool of samples from both domains across features. It was found that all of the $p$ values obtained from independently comparing the feature distributions across domains were zero and that each domains' features were statistically different from one another. 

For clarity it is important to explicitly state the directionality of the hypothesis test that was employed. The hypothesis was posed such that at a significance level of 0.05, for two distributions to be similar, according the Kuiper statistic, a $p$ value should be greater then 0.95. Visually the differences between the quantile trimmed and range normalized feature distributions can be seen in Figure 2.

\begin{figure}[H]
	\centering
	\includegraphics[width=0.70\textwidth]{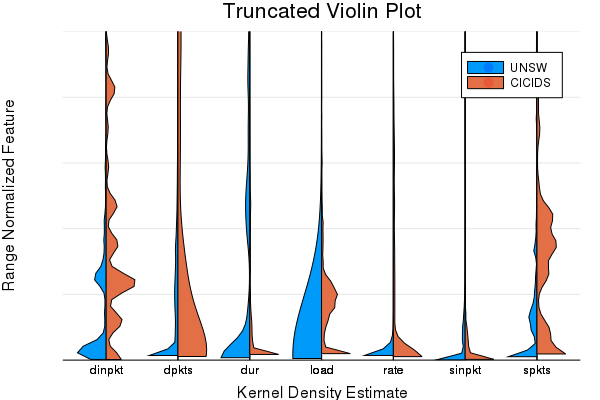}
	\caption{A truncated violin plot of the kernel density estimated feature distributions (bins = 1,000, N = 80,000). The long tails of the distributions were truncated for visualization purposes so that only the smallest 5\% of values are displayed. }
\end{figure}

An empirical test where a K-Nearest Neighbors (K-NN) classification model was transferred from UNSW NB-15 to the CICIDS2017 `Friday' subset. This test was performed to confirm that the inter-domain feature distributions were not likely well represented by one another in terms of magnitude or localization. K-NN was selected because K-NN classification models have the ability to learn challenging nonlinear class boundaries for low dimensional problems and offer the statistical guarantee that the Bayes error rate for K-NN models incorporating only 1 nearest neighbor can not be more than $1/2$ the model error when sufficient samples are present. Most importantly, K-NN classifies entries based on the distance between an unknown point and a memory store of labeled samples. Thus, if the normal and DoS representations were similar across domains then K-NN modeling efficacy should also be transferable across domains to some extent. An approximate ball-tree K-NN algorithm to handle the large number of samples available in either distribution with little consequence to the K-NN classification mechanism \cite{ANN}. The results of this experiment are shown in Table 2.

\begin{table}[H]
	\centering
	\begin{tabular}{ c c c }
		Transfer Method & CICIDS2017 Train & CICIDS2017 Test \\ \hline
		Control & 1.0 & 0.966 \\
		No Transfer & 0.445 & 0.432 \\
		CORAL & 0.566 & 0.578    
	\end{tabular}
	\caption{F-Measures of 1-NN models transferred from UNSW NB-15 to CICIDS2017.}
\end{table}

It was found that neither of the transfers offered results which were nearly as effective as the control (training and testing on CICIDS2017 data). Without adapting the CICIDS2017 domain to that of UNSW NB-15 the 1-NN model afforded F measures which were not satisfactory given the class balance of the CICIDS2017 train (1 : 1.3) and test sets (1 : 1.3). Ultimately, both the no transfer trial and the CORAL transfers can be seen as scoring about as well as the no information rate for either class proportion (1/2.3 and 1.3/2.3). These findings coincide with the hypothesis that the domains have different representations of their class labels despite having semantically similar features. These results were somewhat unexpected. In our hands CORAL has been a competitive domain transfer method for other types of data, and many DoS attack vectors are known present similarly to one another \cite{DosReview}.

The evidence gathered from the aforementioned experiments suggested that transferring DoS attacks from these two academic datasets is not a trivial task because the two domains do not natively share similar class representations. The feature distributions were statistically different, and even with scaled/normalized features the CORAL method failed to produce classification based domain transfers above the no information rate. These results encouraged revisiting the problem statement and focusing on what an ideal result might look like. 

\subsection{Transfer of Triage Models}
What should an analyst experience when they bring a new system online? For an analyst to be most effective the first few items presented in a ticketing system should be of high priority for action \cite{Triage}. This is somewhat contrary to how many machine learning approaches phrase the problem of cyber security as one of classification. Not every classification method readily supports triage. For example, K-NN with 1 nearest neighbor does not directly suggest how well a sample fits into a class label unless a heuristic based on sample-to-class distance is imposed such as what is done in Probabilistic Neural Networks \cite{PNN}. 

A neural network architecture that was readily amenable to triage and adversarial learning methods was investigated. In order to assess the efficacy of triage models generated we introduce a plot called the Rolling TopN Accuracy plot. The plot is generated by sorting the prediction vector in descending order and at every location in the abscissa the number of true malicious entries ($M$) observed is divided by the number of samples scored ($N$) is calculated. This measure can be expressed as the following,
$$Rolling Top_N = \frac{\sum_N{ C } }{ N }, \quad C \in \begin{pmatrix} 1, & \mbox{if Class of Interest} \\ 0, & \mbox{else} \end{pmatrix}$$

\noindent Each point on the plot is a score between 0-1, and the plot describes how often samples labeled to be of high priority (left-most) were actually the correct class label/priority level. For example, if the first 100 highest scoring samples ranked by the model, are not DoS, the scores for the first 100 samples will be 0\%. 

The Rolling TopN analysis was first conducted on 10 replicates of the adversarial transfer and 100 replicates for both control transfer methods from the UNSW NB-15 DoS dataset to the CICIDS2017 "Wednesday" subset. CORAL and training a model without the adversarial component (\textit{Naive}) were again used as a controls. It was found that, the adversarial method outperformed both, especially in the first $\approx 1/3$ of samples analyzed. This finding suggested that the adversarial approach can improve domain representations for transfer learning tasks of net flow data. For triage purposes the Top100 predicted samples were also analyzed (Figure 2). It was found that, although the adversarial model did not, on average, provide high scores with $>50.$\% accuracy the method did, on average, perform better then the lower bound(0.\%); a feat the other methods failed to do. 

\begin{figure}[H]
	\centering
	\includegraphics[width=12cm]{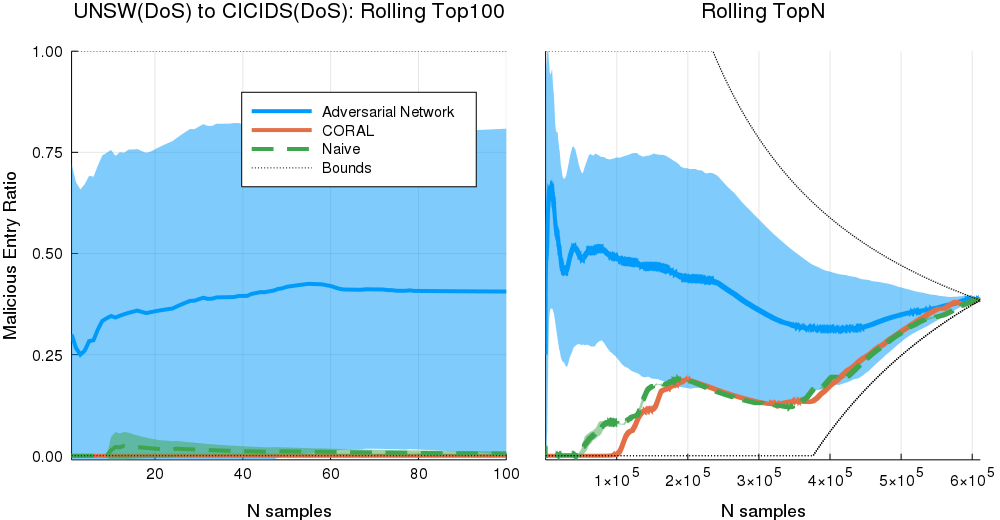} 

	\caption{10 replicates for the Rolling TopN and Top100 Accuracy of the adversarial transfer of UNSW NB-15 DoS to CICIDS2017 "Friday" DoS data. The naive transfer and CORAL transfers were individually performed with 100 replicates. 1 sigma error bars are displayed with a shadow, and the upper and lower bounds for perfect and worse case results are displayed with dashed lines.}
\end{figure}

Next the transfer from CICIDS2017's "Wednesday" DoS subset to UNSW NB-15 DoS was considered. Again, the rolling Top100 accuracies outperformed the control methods (Figure 3). The accuracy across N samples for the control methods were effectively equal to the lower bound for the Top100 samples; meaning they did not, on average, provide usable rankings. This result provided further evidence that suggested the adversarial network approach tended to adapt the class representations across domains. However, the Rolling TopN Accuracy scores going from CICIDS2017's "Wednesday" $\rightarrow$ UNSW NB-15, were comparatively lower than  the transfer from UNSW NB-15 $\rightarrow$ "Wednesday".  

\begin{figure}[H]
	\centering
	\includegraphics[width=12cm]{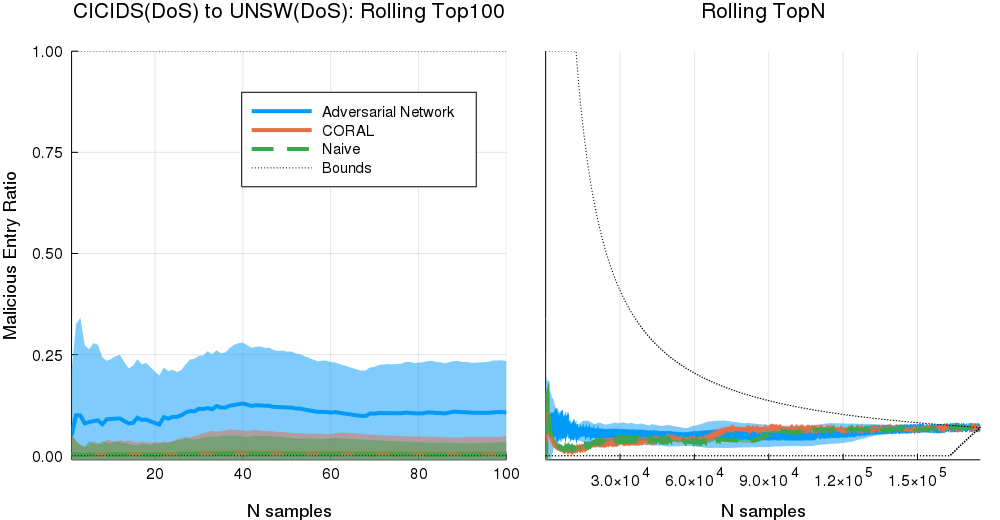} 
	
	\caption{10 replicates for the Rolling TopN and Top100 Accuracy of the transfer trials for transferring CICIDS2017 "Friday" DDoS to UNSW NB-15 DoS attacks. 1 sigma error bars are displayed with a shadow, and the upper and lower bounds for perfect and worse case results are displayed with dashed lines.}
\end{figure}

The success of this transfer lead to questioning whether or not DDoS and DoS detector models could be transferred between one another. To assess this, the transfers both to and from CICIDS2017's "Friday" DDoS subset and the UNSW NB-15 DoS data were considered (Figure 4). Again, the transfer from CICIDS2017 to UNSW-NB 15 tended to have higher scores then the transfer from UNSW-NB 15 to CICIDS2017. Over-all the accuracies obtained from the adversarial transfers using the "Friday" subset tended to be lower in magnitude than that of the "Wednesday" subset. This finding is reasonable, because several types of DoS attacks(volume, protocol, application layer) share similar characteristics with DDoS attacks, however, they are do not necessarily present in the same way because of different vectors. Although the adversarial approach more adequately find invariant representations between both domains than the control methods, it was not without technical shortcomings.

\begin{figure}[H]
	\centering
	\includegraphics[width=12cm]{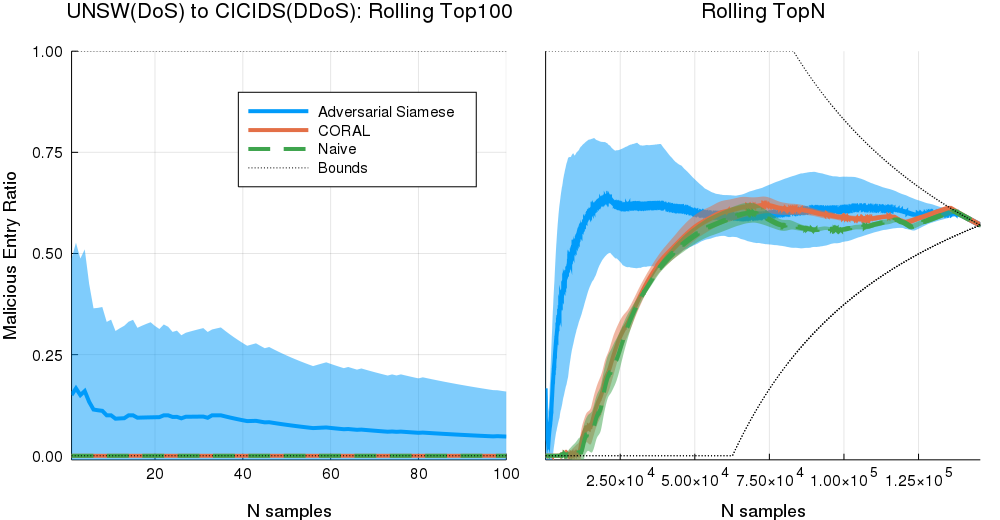}
	\includegraphics[width=12cm]{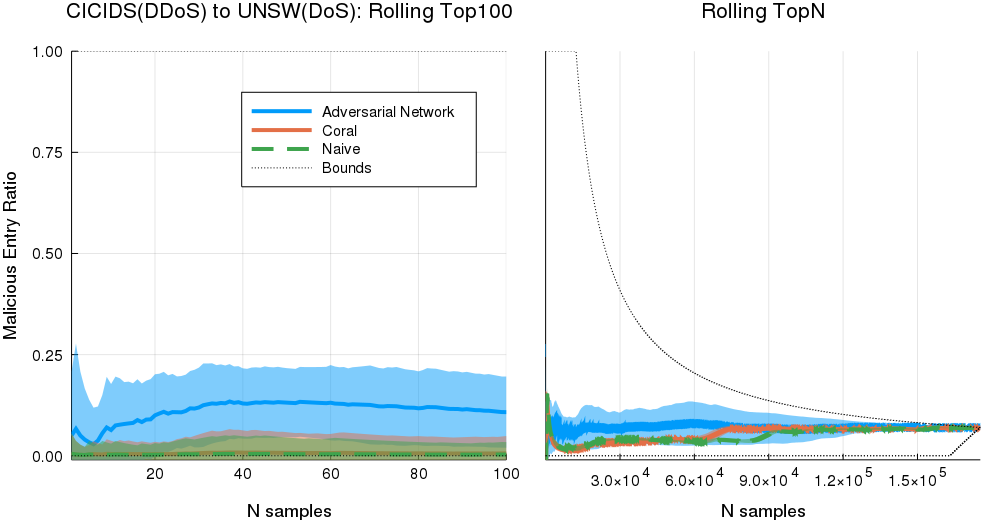}
	
	\caption{10 replicates for the Rolling TopN and Top100 accuracies of the transfer trials for transferring both to and from the CICIDS2017 "Wednesday" DDoS subset and the UNSW NB-15 DoS dataset. 1 sigma error bars are displayed with a shadow, and the upper and lower bounds for perfect and worse case results are displayed with dashed lines.}
\end{figure}

\subsection{Training Considerations}
Training neural networks can be a difficult task because they inherently possess ambiguity in the selection of topologies and values for hyperparameters. Furthermore, training adversarial neural networks adds additional challenges, in that they can be unstable \cite{Stability}, not converge \cite{Converge}, and exhibit otherwise unexpected behavior. Our interests were primarily concerned with adapting domains from a labeled dataset to another dataset which does not have labels. This problem statement, unfortunately, complicated matters further. The question of whether or not a model successfully learned the auxiliary domain becomes pertinent, and to our knowledge, is intractable to answer. 

Despite that the problem has no rigorous solution we found several metrics to be address the efficacy of the model presented. Perhaps most importantly is the question of whether or not the Siamese embedding($E(X)$) is infact making a representation where both domains($A$ \& $B$) are similar statistically. One metric, the congruence measure, describes the difference between the column-wise($u \in {1,2,...,U}$) quantiles of the embedded representations from both domains:

\begin{equation}
	CM_c(X_A, X_B) = \sum_{u=1}^{U} (Pr( X_{A.u} \leq c ) - Pr( X_{B,u} \leq c ) )^ 2
\end{equation}

\noindent By monitoring discrepancies between a suitable range of quantiles(Ie: $c = {0.1,0.2,...,0.9}$) one can determine if over-all, the distributions from $E(X_A)$ or $E(X_B)$ are decreasing(becoming more similar), or if the modeling is driving certain quantiles apart with respect to training iterations. Figure 4 shows are examples of \textit{healthy} and \textit{unhealthy} training plots with respect to 9 quantile differences,

\begin{figure}[H]
	\centering
	\includegraphics[width=10cm]{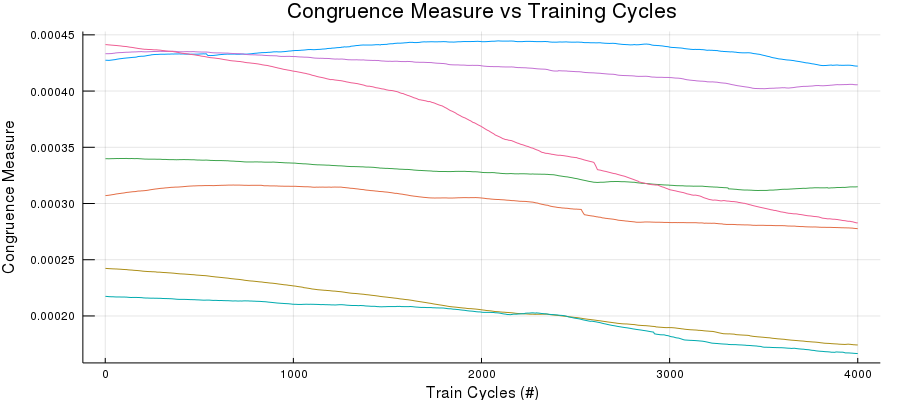} 
	\includegraphics[width=10cm]{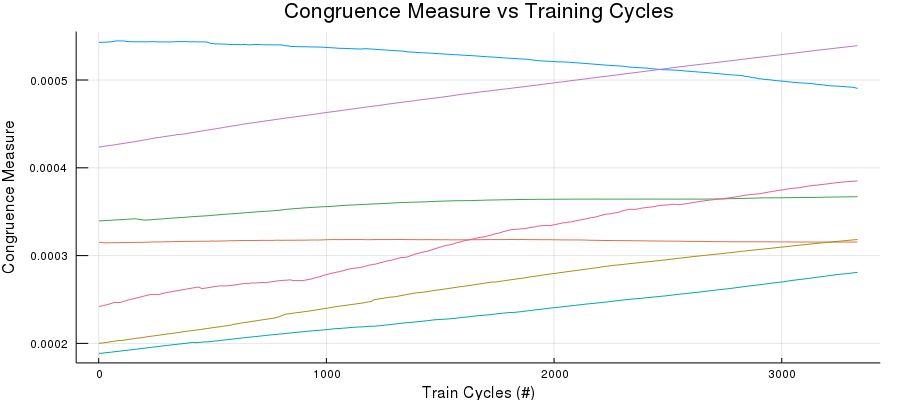}
	
	\caption{Plots of the congruence measure during training for a model that is bringing quantiles closer together(above) and an model which is tending to drive quantiles further apart(below). Each line represents a different quantile. }
\end{figure}

Although the predictive efficacy of the model on the auxiliary domain cannot be directly assessed, another important assumption is that, the learned embedding still performs well on the domain with labels. This can be easily checked by monitoring a TopN plot on the training or a holdout set. Similarly, the typical training statistics (minimum/maximum weight values, ranks of weights, etc) should also be used to ensure the network is \textit{healthy} as usual. One interesting thing to note is that the ListMLE loss can behave erratically when used in an adversarial setting. It is advisable to ensure that the loss follows a smooth pattern and does not have artifacts from local minima. By ensuring that the loss on the supervised domain for the ListMLE subnetwork is providing stable results, and that the quantiles smoothly decrease in value, one can posit that the method performed domain adaptation. It should be stated that with the current task, selected topology, and data used in this study to acquire 10 results that contributed to the Rolling TopN plots, approximately 40 results which did not match these criteria were discarded for either transfer.

\begin{figure}[H]
	\centering
	\includegraphics[width=10cm]{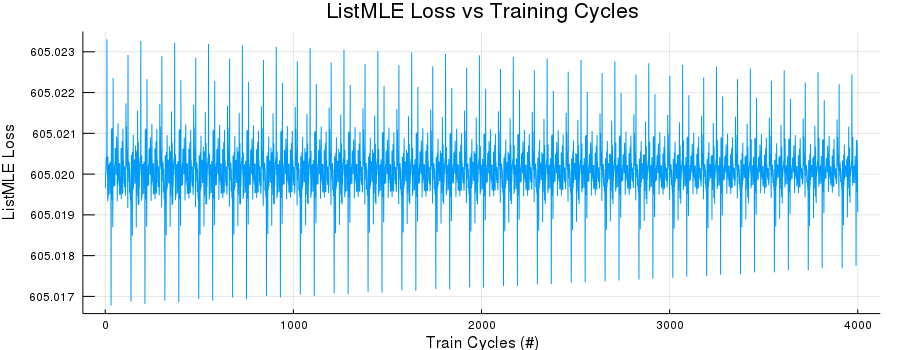} 
	\includegraphics[width=10cm]{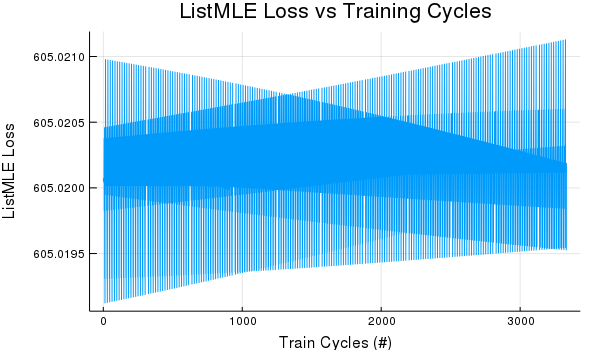}
	
	\caption{Plots of the ListMLE loss during training for the same models shown in Figure 5. The top plot displays a stable loss gradually decaying in variance and the bottom most plot shows a loss plot possibly caught in a local minima where changes in the adversarial layer are leading to unstable behavior (below). }
\end{figure}

\subsection{Conclusion}
It was shown that the UNSW NB-15, and CICIDS2017 datasets have contributions of variance in at least 7 features that make the modeling of DoS attacks difficult to transfer. The problem seemed trivial, but was surprising in how difficult it was to achieve with what was available. To give credit to the difficulty of the problem, it must be considered how much happens in 2 years with regards to attack vectors, internet use trends, and the experimental constraints on these 2 academic datasets. Yet, adversarial network models were shown to improve the representation of latent features in statistically different networking domains for purposes of DoS triage. The semisupervised adversarial network performed better than baseline, and also an established transfer learning method, CORAL. 

Although this study was limited in scope by data availability we wish to discuss how this general approach is not be limited to net-flow like records nor triage tasks. Adversarial neural networks are a general machine learning approach which can be tailored to many applications. However, this work also describes difficulties associated with model tuning and reproducibility of even the simple adversarial neural network topology used. Although this work presents useful training criteria for tuning these models in a semisupervised setting, considerable research can be done to make the training of these models more reliable, and their evaluation less manual. We plan to address some of these issues in future work, but readily invite other researchers to contribute in this field. For global understanding of cyber threats, both generic and distributable representations / understandings must be obtained. If a threat is known by a detector on the public internet, it should be transferable on a private network detectors.

\subsection{Acknowledgements}
The research reported in this document was performed in connection with contract number W911NF-18-C0019 with the U.S. Army Contracting Command - Aberdeen Proving Ground (ACC-APG) and the Defense Advanced Research Projects Agency (DARPA). The views and conclusions contained in this document are those of the authors and should not be interpreted as presenting the official policies or position, either expressed or implied, of ACC-APG, DARPA, or the U.S. Government unless so designated by other authorized documents. Citation of manufacturer’s or trade names does not constitute an official endorsement or approval of the use thereof. The U.S. Government is authorized to reproduce and distribute reprints for Government purposes notwithstanding any copyright notation hereon.



\end{document}